\documentclass[11pt]{article}
\usepackage{acl08}
\usepackage{times}
\usepackage{haldefs}
\usepackage{fancybox}
\usepackage{graphicx}
\usepackage{algorithm}
\usepackage{algorithmic}

\title{Cross-Task Knowledge-Constrained Self Training}

\newcommand{\mytag}[1]{\textsf{\small #1}}
\newcommand{\unlab}{^{\textsf{unlab}}}
\newcommand{\pinch}{\vspace{-2mm}}

\newcommand{\bracket}[2]{\multicolumn{#1}{@{}c@{}}{\textsf{\small [-\hspace{\stretch{1}}#2\hspace{\stretch{1}}-]}}}

\newenvironment{examplebox}%
  {\vspace{2mm}\begin{small}\begin{Sbox}\begin{minipage}{2.8in}\begin{center}\textsc{Running Example}\end{center}}%
  {\end{minipage}\end{Sbox}\end{small}\fbox{\TheSbox}\vspace{2mm}}

\author{Hal Daum\'e III\\School of Computing\\University of Utah\\Salt Lake City, UT 84112\\\url{me@hal3.name}}

\date{}

\begin{document}
\maketitle

\begin{abstract}
  We present an algorithmic framework for learning multiple related
  tasks.  Our framework exploits a form of prior knowledge that relates
  the output spaces of these tasks.  We present PAC learning results
  that analyze the conditions under which such learning is possible.
  We present results on learning a shallow parser and named-entity
  recognition system that exploits our framework, showing consistent
  improvements over baseline methods.
\end{abstract}

\section{Introduction}

When two NLP systems are run on the same data, we expect certain
constraints to hold between their outputs.  This is a form of prior
knowledge.  We propose a self-training framework that uses such
information to significantly boost the performance of one of the
systems.  The key idea is to perform self-training \emph{only} on
outputs that obey the constraints.

Our motivating example in this paper is the task pair: named entity
recognition (NER) and shallow parsing (aka syntactic chunking).
Consider a hidden sentence with known POS and syntactic structure
below.  Further consider four potential NER sequences for this
sentence.

\vspace{2mm}
\begin{small}
%\noindent
\begin{tabular}{@{}r@{ }c@{~~~}c@{~~~}c@{~~~}c@{~~~}c@{~~~}c@{~~~}c@{}}
%George &Bush  &spoke &to    &Congress &yesterday \\
\hline
{\bf POS:}   & \mytag{NNP    }&\mytag{NNP   }&\mytag{VBD   }&\mytag{TO    }&\mytag{NNP      }&\mytag{NN        }\\
%{\bf Chunk:} & \mytag{B-NP   }&\mytag{I-NP  }&\mytag{B-VP  }&\mytag{B-PP  }&\mytag{B-NP     }&\mytag{B-NP      }\\
{\bf Chunk:} & \bracket{2}{NP} & \bracket{1}{VP} & \bracket{1}{PP} & \bracket{1}{NP} & \bracket{1}{NP}
\\
\hline
{\bf NER1:}  & \bracket{3}{Per} & \bracket{1}{O} & \bracket{1}{Org} & \bracket{1}{0} \\
{\bf NER2:}  & \bracket{2}{Per} & \bracket{1}{O} & \bracket{1}{O} & \bracket{1}{O} & \bracket{1}{O} \\
{\bf NER3:}  & \bracket{2}{Per} & \bracket{1}{O} & \bracket{1}{O} & \bracket{2}{Org} \\
{\bf NER4:}  & \bracket{2}{Per} & \bracket{1}{O} & \bracket{1}{O} & \bracket{1}{Org} & \bracket{1}{O} \\
%{\bf NER2:}  & \mytag{B-Per  }&\mytag{I-Per }&\mytag{O     }&\mytag{O     }&\mytag{O        }&\mytag{O         }\\
%{\bf NER3:}  & \mytag{B-Per  }&\mytag{I-Per }&\mytag{O     }&\mytag{O     }&\mytag{B-Org    }&\mytag{I-Org     }\\
%{\bf NER4:}  & \mytag{B-Per  }&\mytag{I-Per }&\mytag{O     }&\mytag{O     }&\mytag{B-Org    }&\mytag{O         }\\
\hline
\end{tabular}
\end{small}
\vspace{2mm}

Without ever seeing the actual sentence, can we guess which NER
sequence is correct?  NER1 seems wrong because we feel like named
entities should not be part of verb phrases.  NER2 seems wrong because
there is an NNP\footnote{When we refer to \mytag{\scriptsize NNP}, we
  also include \mytag{\scriptsize NNPS}.} (proper noun) that is not
part of a named entity (word 5).  NER3 is amiss because we feel it is
unlikely that a \emph{single} name should span more than one NP (last
two words).  NER4 has none of these problems and seems quite
reasonable.  In fact, for the hidden sentence, NER4 is
correct\footnote{The sentence is: ``George Bush spoke to Congress
  today''}.

The remainder of this paper deals with the problem of formulating such
prior knowledge into a workable system.  There are similarities
between our proposed model and both self-training and co-training;
background is given in Section~\ref{sec:background}.  We present a
formal model for our approach and perform a simple, yet informative,
analysis (Section~\ref{sec:model}).  This analysis allows us to define
what good and bad constraints are.  Throughout, we use a running
example of NER using hidden Markov models to show the efficacy of the
method and the relationship between the theory and the implementation.
Finally, we present full-blown results on seven different NER data
sets (one from CoNLL, six from ACE), comparing our method to several
competitive baselines (Section~\ref{sec:experiments}).  We see that
for many of these data sets, less than one hundred labeled NER
sentences are required to get state-of-the-art performance, using a
discriminative sequence labeling algorithm \cite{daume05laso}.

\section{Background} \label{sec:background}

Self-training works by learning a model on a small amount of labeled
data.  This model is then evaluated on a large amount of unlabeled
data.  Its predictions are assumed to be correct, and it is retrained
on the unlabeled data according to its own predictions.  Although
there is little theoretical support for self-training, it is
relatively popular in the natural language processing community.  Its
success stories range from parsing \cite{mcclosky06selftraining} to
machine translation \cite{ueffing06selftraining}.  In some cases,
self-training takes into account \emph{model confidence}.

Co-training \cite{yarowsky95wsd,blum98cotraining} is related to
self-training, in that an algorithm is trained on its own predictions.
Where it differs is that co-training learns two \emph{separate} models
(which are typically assumed to be independent; for instance by
training with disjoint feature sets).  These models are both applied
to a large repository of unlabeled data.  Examples on which these two
models \emph{agree} are extracted and treated as labeled for a new
round of training.  In practice, one often also uses a notion of model
confidence and only extracts agreed-upon examples for which both
models are confident.  The original, and simplest analysis of
co-training is due to Blum and Mitchell \shortcite{blum98cotraining}.
It does not take into account confidence (to do so requires a
\emph{significantly} more detailed analysis
\cite{dasgupta01cotraining}), but is useful for understanding the
process.

\section{Model} \label{sec:model}

We define a formal PAC-style \cite{valiant84learnable} model that we
call the ``hints model''\footnote{The name comes from thinking of our
  knowledge-based constraints as ``hints'' to a learner as
  to what it should do.}.  We have an instance space $\cX$ and
\emph{two} output spaces $\cY_1$ and $\cY_2$.  We assume two concept
classes $\cC_1$ and $\cC_2$ for each output space respectively.  Let
$\cD$ be a distribution over $\cX$, and $f_1 \in \cC_1$ (resp., $f_2
\in \cC_2$) be target functions.  The goal, of course, is to use a
finite sample of examples drawn from $\cD$ (and labeled---perhaps with
noise---by $f_1$ and $f_2$) to ``learn'' $h_1 \in \cC_1$ and $h_2 \in
\cC_2$, which are good approximations to $f_1$ and $f_2$.

So far we have not made use of any notion of constraints.  Our
expectation is that if we constrain $h_1$ and $h_2$ to \emph{agree}
(vis-a-vis the example in the Introduction), then we should need fewer
labeled examples to learn either.  (The agreement should ``shrink''
the size of the corresponding hypothesis spaces.)  To formalize this,
let $\chi : \cY_1 \times \cY_2 \fto \{0,1\}$ be a \emph{constraint
  function.}  We say that two outputs $y_1 \in \cY_1$ and $y_2 \in
\cY_2$ are \emph{compatible} if $\chi(y_1,y_2) = 1$.  We need to
assume that $\chi$ is correct:

\pinch
\begin{mydefinition}
  We say that $\chi$ is \emph{correct} with respect to $\cD, f_1, f_2$
  if whenever $x$ has non-zero probability under $\cD$, then
  $\chi(f_1(x),f_2(x)) = 1$.
\end{mydefinition}
\pinch

\begin{examplebox}
  In our example, $\cY_1$ is the space of all POS/chunk sequences and
  $\cY_2$ is the space of all NER sequences.  We assume that $\cC_1$
  and $\cC_2$ are both represented by HMMs over the appropriate state
  spaces.  The functions we are trying to learn are $f_1$, the
  ``true'' POS/chunk labeler and $f_2$, the ``true'' NER labeler.
  (Note that we assume $f_1 \in \cC_1$, which is obviously not true
  for language.)\\

  Our constraint function $\chi$ will require the following for
  agreement: (1) any NNP must be part of a named entity; (2) any named
  entity must be a subsequence of a noun phrase.  This is precisely
  the set of constraints discussed in the introduction.
\end{examplebox}

The question is: given this additional source of knowledge (i.e.,
$\chi$), has the learning problem become easier?  That is, can we
learn $f_2$ (and/or $f_1$) using significantly fewer labeled examples
than if we did not have $\chi$?  Moreover, we have assumed that $\chi$
is \emph{correct}, but is this enough?  Intuitively, no: a function
$\chi$ that returns $1$ regardless of its inputs is clearly not
useful.  Given this, what other constraints must be placed on $\chi$.
We address these questions in Sections~\ref{sec:model:analysis}.
However, first we define our algorithm.

\subsection{One-sided Learning with Hints} \label{sec:model:algorithm}

We begin by considering a simplified version of the ``learning with
hints'' problem.  Suppose that all we care about is learning $f_2$.
We have a small amount of data labeled by $f_2$ (call this $D$) and a
\emph{large} amount of data labeled by $f_1$ (call this
$D\unlab$--''unlab'' because as far as $f_2$ is concerned, it is
unlabeled).

\begin{examplebox}
  In our example, this means that we have a small amount of labeled
  NER data and a large amount of labeled POS/chunk data.  We use
  $3500$ sentences from CoNLL \cite{sang03conll-ne} as the NER data
  and section 20-23 of the WSJ
  \cite{marcus93treebank,ramshaw95chunking} as the POS/chunk data
  ($8936$ sentences).  We are \emph{only} interested in learning to do
  NER.  Details of the exact HMM setup are in
  Section~\ref{sec:hmm}.
\end{examplebox}

We call the following algorithm ``One-Sided Learning with Hints,''
since it aims only to learn $f_2$:

\begin{algorithmic}[1]
\STATE Learn $h_2$ directly on $D$
\STATE For each example $(x,y_1) \in D\unlab$
\STATE ~~~~Compute $y_2 = h_2(x)$
\STATE ~~~~If $\chi(y_1,y_2)$, add $(x,y_2)$ to $D$
\STATE Relearn $h_2$ on the (augmented) $D$
\STATE Go to (2) if desired
\end{algorithmic}

\begin{examplebox}
  In step 1, we train an NER HMM on CoNLL.  On test data, this model
  achieves an $F$-score of $50.8$.  In step 2, we run this HMM on all
  the WSJ data, and extract $3145$ compatible examples.  In step 3, we
  retrain the HMM; the $F$-score rises to $58.9$.
\end{examplebox}

\subsection{Two-sided Learning with Hints} \label{sec:twos}

In the two-sided version, we assume that we have a small amount of
data labeled by $f_1$ (call this $D_1$), a small amount of data
labeled by $f_2$ (call this $D_2$) and a \emph{large} amount of
unlabeled data (call this $D\unlab$).  The algorithm we propose for
learning hypotheses for both tasks is below:

\begin{algorithmic}[1]
\STATE Learn $h_1$ on $D_1$ and $h_2$ on $D_2$.
\STATE For each example $x \in D\unlab$:
\STATE ~~~~Compute $y_1 = h_1(x)$ and $y_2 = h_2(x)$
\STATE ~~~~If $\chi(y_1,y_2)$ add $(x,y_1)$ to $D_1$, $(x,y_2)$ to $D_2$ 
\STATE Relearn $h_1$ on $D_1$ and $h_2$ on $D_2$.
\STATE Go to (2) if desired
\end{algorithmic}

\begin{examplebox}
  We use $3500$ examples from NER and $1000$ from WSJ.  We use the
  remaining $18447$ examples as unlabeled data.  The baseline HMMs
  achieve $F$-scores of $50.8$ and $76.3$, respectively.  In step 2,
  we add $7512$ examples to each data set.  After step 3, the new
  models achieve $F$-scores of $54.6$ and $79.2$, respectively.  The
  gain for NER is lower than before as it is trained against ``noisy''
  syntactic labels.
\end{examplebox}

\subsection{Analysis} \label{sec:model:analysis}

Our goal is to prove that one-sided learning with hints ``works.''
That is, if $C_2$ is learnable from large amounts of labeled data,
then it is also learnable from small amounts of labeled data and large
amounts of $f_1$-labeled data.  This is formalized in
Theorem~\ref{thm:pac} (all proofs are in Appendix~\ref{sec:proofs}).
However, before stating the theorem, we must define an ``initial
weakly-useful predictor'' (terminology from Blum and
Mitchell\shortcite{blum98cotraining}), and the notion of noisy
PAC-learning in the structured domain.

\begin{mydefinition}
We say that $h$ is a \emph{weakly-useful predictor} of $f$ if for all $y$:
%\begin{small}
%\begin{enumerate}
$\Pr_{\cD}\left[ h(x) = y \right] \geq \ep$
and $\Pr_{\cD}\left[ f(x) = y \| h(x) = y' \neq y \right] \geq \Pr_{\cD}\left[ f(x) = y \right] + \ep$.
%\end{enumerate}
%\end{small}
\end{mydefinition}

This definition simply ensures that (1) $h$ is nontrivial: it assigns
some non-zero probability to every possible output; and (2) $h$ is
somewhat indicative of $f$.  In practice, we use the hypothesis
learned on the small amount of training data during step (1) of the
algorithm as the weakly useful predictor.

\begin{mydefinition}
  We say that $\cC$ is \emph{PAC-learnable with noise (in the
    structured setting)} if there exists an algorithm with the
  following properties.  For any $c \in \cC$, any distribution $\cD$
  over $\cX$, any $0 \leq \eta \leq 1/\card{\cY}$, any $0<\ep<1$, any
  $0 < \de< 1$ and any $\eta \leq \eta_0 < 1/\card{\cY}$, if the
  algorithm is given access to examples drawn
  $\text{EX}^\et_{SN}(c,\cD)$ and inputs $\ep,\de$ and $\eta_0$, then
  with probability at least $1-\de$, the algorithm returns a
  hypothesis $h \in \cC$ with error at most $\ep$.  Here,
  $\text{EX}^\et_{SN}(c,\cD)$ is a structured noise oracle, which
  draws examples from $\cD$, labels them by $c$ and randomly replaces
  with another label with prob. $\eta$.
\end{mydefinition}

Note here the rather weak notion of noise: entire structures are
randomly changed, rather than individual labels.  Furthermore, the
error is 0/1 loss over the entire structure.  Collins
\shortcite{collins01estimation} establishes learnability results for
the class of hyperplane models under 0/1 loss.  While not stated
directly in terms of PAC learnability, it is clear that his results
apply.  Taskar et al. \shortcite{taskar05mmmn} establish
\emph{tighter} bounds for the case of Hamming loss.  This suggests
that the requirement of 0/1 loss is weaker.

As suggested before, it is not sufficient for $\chi$ to simply be
\emph{correct} (the constant $1$ function is correct, but not useful).
We need it to be discriminating, made precise in the following
definition.

\begin{mydefinition} \label{def:discriminating} We say the
  \emph{discrimination} of $\chi$ for $h^0$ is $\Pr_\cD[ \chi(f_1(x),
  h^0(x)) ]\inv$.
\end{mydefinition}

In other words, a constraint function is discriminating when it is
unlikely that our weakly-useful predictor $h^0$ chooses an output that
satisfies the constraint.  This means that if we \emph{do} find
examples (in our unlabeled corpus) that satisfy the constraints, they
are likely to be ``useful'' to learning.

\begin{examplebox}
  In the NER HMM, let $h^0$ be the HMM obtained by training on the
  small labeled NER data set and $f_1$ is the true syntactic labels.
  We approximate $\Pr_\cD$ by an empirical estimate over the unlabeled
  distribution.  This gives a discrimination is $41.6$ for the
  constraint function defined previously.  However, if we compare
  against ``weaker'' constraint functions, we see the appropriate
  trend.  The value for the constraint based only on POS tags is
  $39.1$ (worse) and for the NP constraint alone is $27.0$ (much
  worse).
\end{examplebox}  

\begin{mytheorem} \label{thm:pac} Suppose $C_2$ is PAC-learnable with
  noise in the structured setting, $h^0_2$ is a weakly useful
  predictor of $f_2$, and $\chi$ is correct with respect to
  $\cD,f_1,f_2,h^0_2$, and has discrimination $\geq
  2(\card{\cY}-1)$.  Then $C_2$ is also PAC-learnable with one-sided
  hints.
\end{mytheorem}

The way to interpret this theorem is that it tells us that if the
initial $h_2$ we learn in step 1 of the one-sided algorithm is ``good
enough'' (in the sense that it is weakly-useful), then we can use it
as specified by the remainder of the one-sided algorithm to obtain an
arbitrarily good $h_2$ (via iterating).

The dependence on $\card{\cY}$ is the discrimination bound for $\chi$
is unpleasant for structured problems.  If we wish to find $M$
unlabeled examples that satisfy the hints, we'll need a total of at
least $2M(\card{\cY}-1)$ total.  This dependence can be improved as
follows.  Suppose that our structure is represented by a graph over
vertices $V$, each of which can take a label from a set $Y$.  Then,
$\card{\cY} = \card{Y^V}$, and our result requires that $\chi$ be
discriminating on an order exponential in $V$.  Under the assumption
that $\chi$ decomposes over the graph structure (true for our example)
and that $C_2$ is PAC-learnable with per-vertex noise, then the
discrimination requirement drops to $2\card{V}(\card{Y}-1)$.

\begin{examplebox}
  In NER, $\card{Y}=9$ and $\card{V}\approx 26$.  This means that the
  values from the previous example look not quite so bad.  In the 0/1
  loss case, they are compared to $10^{25}$; in
  the Hamming case, they are compared to only $416$.
  The ability of the one-sided algorithm follows the same trends as
  the discrimination values.  Recall the baseline performance is
  $50.8$.  With both constraints (and a discrimination value of
  $41.6$), we obtain a score of $58.9$.  With just the POS constraint
  (discrimination of $39.1$), we obtain a score of $58.1$.  With just
  the NP constraint (discrimination of $27.0$, we obtain a score of
  $54.5$.
\end{examplebox}

The final question is how one-sided learning relates to two-sided
learning.  The following definition and easy corollary shows that they
are related in the obvious manner, but depends on a notion of
uncorrelation between $h^0_1$ and $h^0_2$.

\begin{mydefinition}
  We say that $h_1$ and $h_2$ are \emph{uncorrelated}
  if $\Pr_{\cD} \left[ h_1(x) = y_1 \| h_2(x) = y_2, x \right] =
  \Pr_{\cD} \left[ h_1(x) = y_1 \| x \right]$.
\end{mydefinition}

\begin{mycorollary} \label{cor:twosided} Suppose $C_1$ and $C_2$ are
  both PAC-learnable in the structured setting, $h^0_1$ and $h^0_2$
  are weakly useful predictors of $f_1$ and $f_2$, and $\chi$ is
  correct with respect to $\cD,f_1,f_2,h^0_1$ and $h^0_2$, and has
  discrimination $\geq 4(\card{\cY}-1)^2$ (for 0/1 loss) or $\geq
  4\card{V}^2(\card{Y}-1)^2$ (for Hamming loss), and that $h^0_1$ and
  $h^0_2$ are uncorrelated.  Then $C_1$ and $C_2$ are also
  PAC-learnable with two-sided hints.
\end{mycorollary}

Unfortunately, Corollary~\ref{cor:twosided} depends
\emph{quadratically} on the discrimination term, unlike
Theorem~\ref{thm:pac}.

\section{Experiments} \label{sec:experiments}

In this section, we describe our experimental results.  We have
already discussed some of them in the context of the running example.
In Section~\ref{sec:data}, we briefly describe the data sets we use.
A full description of the HMM implementation and its results are in
Section~\ref{sec:hmm}.  Finally, in Section~\ref{sec:laso}, we present
results based on a competitive, discriminatively-learned sequence
labeling algorithm.  All results for NER and chunking are in terms of
F-score; all results for POS tagging are accuracy.

\subsection{Data Sets} \label{sec:data}

Our results are based on syntactic data drawn from the Penn Treebank
\cite{marcus93treebank}, specifically the portion used by CoNLL 2000
shared task \cite{sang00chunking}.  Our NER data is from two sources.
The first source is the CoNLL 2003 shared task date
\cite{sang03conll-ne} and the second source is the 2004 NIST Automatic
Content Extraction \cite{ace04}.  The ACE data constitute six separate
data sets from six domains: weblogs (wl), newswire (nw), broadcast
conversations (bc), United Nations (un), direct telephone speech (dts)
and broadcast news (bn).  Of these, bc, dts and bn are all speech data
sets.  All the examples from the previous sections have been limited
to the CoNLL data.

\subsection{HMM Results} \label{sec:hmm}

The experiments discussed in the preceding sections are based on a
generative hidden Markov model for both the NER and syntactic
chunking/POS tagging tasks.  The HMMs constructed use first-order
transitions and emissions.  The emission vocabulary is pruned so that
any word that appears $\leq 1$ time in the training data is replaced
by a unique {\tt *unknown*} token.  The transition and emission
probabilities are smoothed with Dirichlet smoothing, $\al = 0.001$
(this was not-aggressively tuned by hand on one setting).  The HMMs
are implemented as finite state models in the Carmel toolkit
\cite{carmel}.

The various compatibility functions are also implemented as finite
state models.  We implement them as a transducer \emph{from} POS/chunk
labels \emph{to} NER labels (though through the reverse operation,
they can obviously be run in the opposite direction).  The
construction is with a single state with transitions:
\begin{itemize}
\pinch\pinch\item \mytag{(NNP,?)} maps to \mytag{B-*} and \mytag{I-*}
\pinch\pinch\item \mytag{(?,B-NP)} maps to \mytag{B-*} and \mytag{O}
\pinch\pinch\item \mytag{(?,I-NP)} maps to \mytag{B-*}, \mytag{I-*} and \mytag{O}
\pinch\pinch\item Single exception: \mytag{(NNP,x)}, where \mytag{x} is \emph{not}
  an NP tag maps to anything (this is simply to avoid empty
  composition problems).  This occurs in $100$ of the $212k$ words in
  the Treebank data and more rarely in the automatically tagged
  data.
\end{itemize}

\subsection{One-sided Discriminative Learning} \label{sec:laso}

\begin{figure*}
\includegraphics[width=\textwidth]{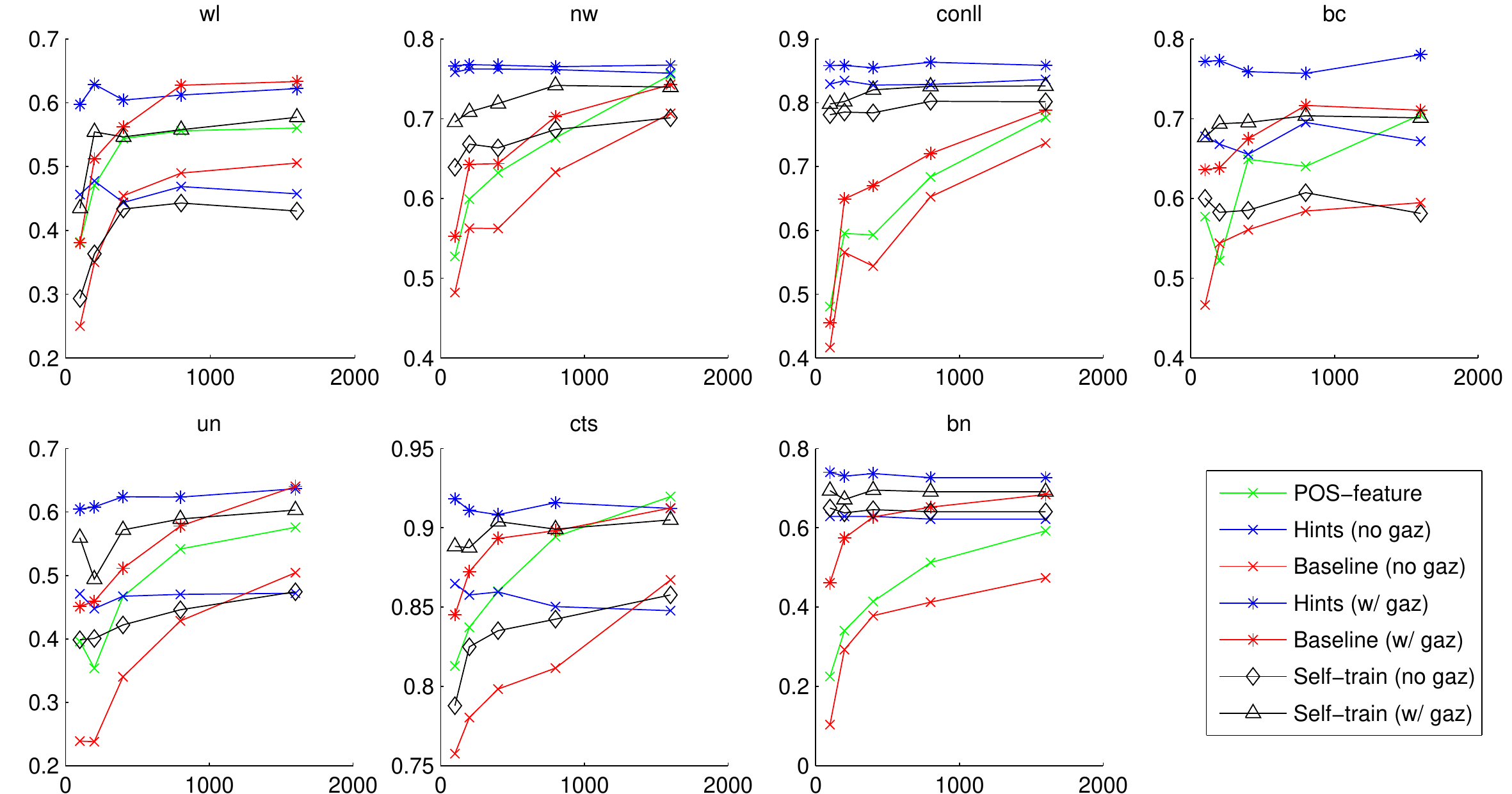}
\caption{Results of varying the amount of NER labeled data, for a
  fixed amount ($M=8936$) of syntactic data.}
\label{fig:vary-labeled}
\end{figure*}

In this section, we describe the results of one-sided discriminative
labeling with hints.  We use the true syntactic labels from the Penn
Treebank to derive the constraints (this is roughly $9000$ sentences).
We use the LaSO sequence labeling software \cite{daume05laso}, with
its built-in feature set.

Our goal is to analyze two things: (1) what is the effect of the
amount of labeled NER data?  (2) what is the effect of the amount of
labeled syntactic data from which the hints are constructed?

To answer the first question, we keep the amount of syntactic data
fixed (at $8936$ sentences) and vary the amount of NER data in $N \in
\{100, 200, 400, 800, 1600\}$.  We compare models with and without the
default gazetteer information from the LaSO software.  We have the
following models for comparison:

\begin{itemize}
\pinch\pinch\item A default ``Baseline'' in which we simply train
  the NER model without using syntax.

\pinch\pinch\item In ``POS-feature'', we do the same thing, but we first label the
  NER data using a tagger/chunker trained on the $8936$ syntactic
  sentences.  These labels are used as features for the baseline.

\pinch\pinch\item A ``Self-training'' setting where we use the $8936$ syntactic
  sentences as ``unlabeled,'' label them with our model, and then
  train on the results.  (This is equivalent to a hints model where
  $\chi(\cdot,\cdot)=1$ is the constant $1$ function.)  We use model
  confidence as in Blum and
  Mitchell~\shortcite{blum98cotraining}.\footnote{Results without
    confidence were significantly worse.}
\end{itemize}
\pinch

The results are shown in Figure~\ref{fig:vary-labeled}.  The trends we
see are the following:

\begin{itemize}
\pinch\pinch\item More data always helps.
\pinch\pinch\item Self-training usually helps over the baseline  (though not always: for instance in wl
  and parts of cts and bn).
\pinch\pinch\item Adding the gazetteers help.
\pinch\pinch\item Adding the syntactic features helps.
\pinch\pinch\item Learning with hints, especially for $\leq 1000$ training data
  points, helps significantly, even over self-training.
\end{itemize}
\pinch

We further compare the algorithms by looking at how many training
setting has each as the winner.  In particular, we compare both hints
and self-training to the two baselines, and then compare hints to
self-training.  If results are not significant at the 95\% level
(according to McNemar's test), we call it a tie.  The results are in
Table~\ref{tab:vary-labeled}.

\begin{table}[t]
\center
\begin{tabular}{|r|cc|c|}
\hline
& {\bf Hints} & {\bf Self-T} & {\bf Hints} \\
& vs {\bf Base} & vs {\bf Base} & vs {\bf Self-T} \\
\hline
{\bf Win}  & $29$ & $20$ & $24$ \\
{\bf Tie}  & $6$ & $12$ & $11$ \\
{\bf Lose} & $0$ & $3$ & $0$ \\
\hline
\end{tabular}
\caption{Comparison between hints, self-training and the (best)
  baseline for varying amount of labeled data.}
\label{tab:vary-labeled}
\end{table}

\begin{figure*}
\includegraphics[width=\textwidth]{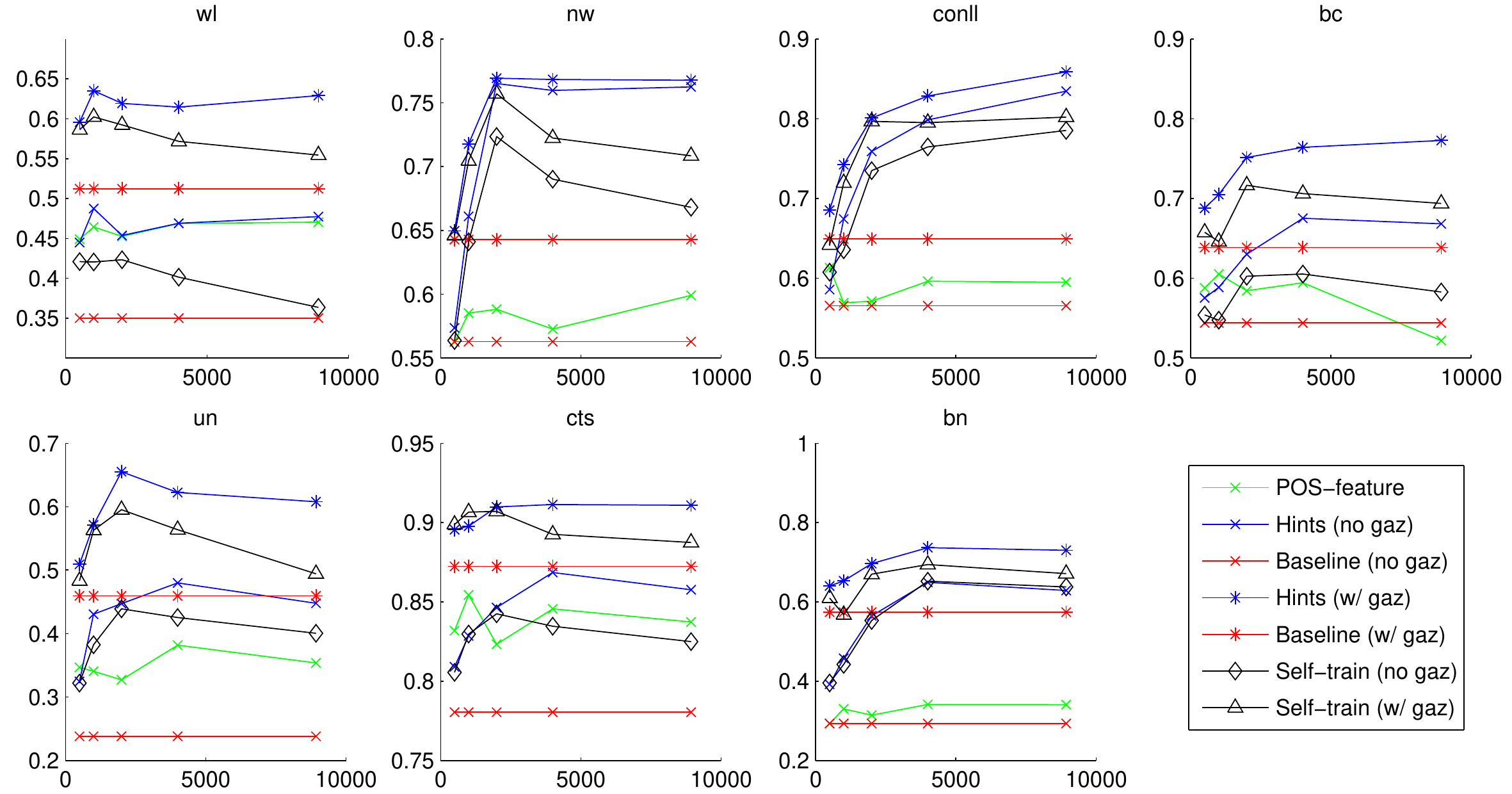}
\caption{Results of varying amount of syntactic data for a fixed
  amount of NER data ($N=200$ sentences).}
\label{fig:vary-unlabeled}
\end{figure*}

In our second set of experiments, we consider the role of the
syntactic data.  For this experiment, we hold the number of NER
labeled sentences constant (at $N=200$) and vary the amount of
syntactic data in $M \in \{ 500, 1000, 2000, 4000, 8936 \}$.  The
results of these experiments are in Figure~\ref{fig:vary-unlabeled}.
The trends are:

\begin{itemize}
\pinch\item The POS feature is relatively insensitive to the amount of
  syntactic data---this is most likely because it's weight is
  discriminatively adjusted by LaSO so that if the syntactic
  information is bad, it is relatively ignored.
\pinch\item Self-training performance often \emph{degrades} as the amount of
  syntactic data increases.
\pinch\item The performance of learning with hints increases steadily with
  more syntactic data.
\end{itemize}

As before, we compare performance between the different models,
declaring a ``tie'' if the difference is not statistically significant
at the 95\% level.  The results are in Table~\ref{tab:vary-unlabeled}.

\begin{table}[t]
\center
\begin{tabular}{|r|cc|c|}
\hline
& {\bf Hints} & {\bf Self-T} & {\bf Hints} \\
& vs {\bf Base} & vs {\bf Base} & vs {\bf Self-T} \\
\hline
{\bf Win}  & $34$ & $28$ & $15$ \\
{\bf Tie}  & $1$  & $7$  & $20$ \\
{\bf Lose} & $0$  & $0$  & $0$ \\
\hline
\end{tabular}
\caption{Comparison between hints, self-training and the (best)
  baseline for varying amount of unlabeled data.}
\label{tab:vary-unlabeled}
\end{table}

In experiments not reported here to save space, we experimented with
several additional settings.  In one, we weight the unlabeled data in
various ways: (1) to make it equal-weight to the labeled data; (2) at
$10\%$ weight; (3) according to the score produced by the first round
of labeling.  None of these had a significant impact on scores; in a
few cases performance went up by $\ll 1$, in a few cases, performance
went down about the same amount.

\subsection{Two-sided Discriminative Learning} \label{sec:two-sided}

In this section, we explore the use of two-sided discriminative
learning to boost the performance of our syntactic chunking, part of
speech tagging, and named-entity recognition software.  We continue to
use LaSO \cite{daume05laso} as the sequence labeling technique.

The results we present are based on attempting to improve the
performance of a state-of-the-art system train on \emph{all} of the
training data.  (This is in contrast to the results in
Section~\ref{sec:laso}, in which the effect of using limited amounts
of data was explored.)  For the POS tagging and syntactic chunking, we
being with all $8936$ sentences of training data from CoNLL.  For the
named entity recognition, we limit our presentation to results from
the CoNLL 2003 NER shared task.  For this data, we have roughly $14k$
sentences of training data, all of which are used.  In both cases, we
reserve $10\%$ as development data.  The development data is use to do
early stopping in LaSO.

As unlabeled data, we use $1m$ sentences extracted from the North
American National Corpus of English  (previously used for
self-training of parsers \cite{mcclosky06selftraining}).  These $1m$
sentences were selected by dev-set relativization against the union of
the two development data sets.

Following similar ideas to those presented by Blum and
Mitchell~\shortcite{blum98cotraining}, we employ two slight
modifications to the algorithm presented in Section~\ref{sec:twos}.
First, in step (2b) instead of adding \emph{all} allowable instances
to the labeled data set, we only add the top $R$ (for some
hyper-parameter $R$), where ``top'' is determined by average model
confidence for the two tasks.  Second, Instead of using the
\emph{full} unlabeled set to label at each iteration, we begin with
a random subset of $10R$ unlabeled examples and another add random
$10R$ every iteration.

We use the same baseline systems as in one-sided learning: a Baseline
that learns the two tasks independently; a variant of the Baseline on
which the output of the POS/chunker is used as a feature for the NER;
a variant based on self-training; the hints-based method.  In all
cases, we \emph{do} use gazetteers.  We run the hints-based model for
$10$ iterations.  For self-training, we use $10R$ unlabeled examples
(so that it had access to the same amount of unlabeled data as the
hints-based learning after all $10$ iterations).  We used three values
of $R$: $50$, $100$, $500$.  We select the best-performing model (by
the dev data) over these ten iterations.  The results are in
Table~\ref{tab:summary}.

\begin{table}[t]
\center
\begin{tabular}{|l|c|c|}
\hline
& {\bf Chunking} & {\bf NER} \\
\hline
{\bf Baseline} & 94.2  &  87.5 \\
\hline
{\bf w/POS} & N/A & 88.0 \\
\hline
{\bf Self-train} & & \\
$R=50 $ & 94.2 & 88.0 \\
$R=100$ & 94.3 & 88.6 \\
$R=500$ & 94.1 & 88.4 \\
\hline
{\bf Hints} &  &  \\
$R=50 $ & 94.2 & 88.5 \\
$R=100$ & 94.3 & 89.1 \\
$R=500$ & 94.3 & 89.0 \\
\hline
\end{tabular}
\caption{Results on two-sided learning with hints.}
\label{tab:summary}
\end{table}

As we can see, performance for syntactic chunking is relatively
stagnant: there are no significant improvements for any of the methods
over the baseline.  This is not surprising: the form of the constraint
function we use tells us a \emph{lot} about the NER task, but
relatively little about the syntactic chunking task.  In particular,
it tells us nothing about phrases other than NPs.  On the other hand,
for NER, we see that both self-training and learning with hints
improve over the baseline.  The improvements are not enormous, but
\emph{are} significant (at the 95\% level, as measured by McNemar's
test).  Unfortunately, the improvements for learning with hints over
the self-training model are only significant at the 90\% level.

\section{Discussion} \label{sec:discussion}

We have presented a method for simultaneously learning two tasks using
prior knowledge about the relationship between their outputs.  This is
related to \emph{joint inference} \cite{jointinference}.  However, we
do not require that that a single data set be labeled for multiple
tasks.  In all our examples, we use separate data sets for shallow
parsing as for named-entity recognition.  Although all our experiments
used the LaSO framework for sequence labeling, there is \emph{noting}
in our method that assumes any particular learner; alternatives
include: conditional random fields~\cite{lafferty01crf}, independent
predictors~\cite{punyakanok01inference}, max-margin Markov
networks~\cite{taskar05mmmn}, etc.

Our approach, both algorithmically and theoretically, is most related
to ideas in co-training \cite{blum98cotraining}.  The key difference
is that in co-training, one assumes that the two ``views'' are on the
\emph{inputs}; here, we can think of the two output spaces as being
the difference ``views'' and the compatibility function $\chi$ being a
method for reconciling these two views.  Like the pioneering work of
Blum and Mitchell, the algorithm we employ in practice makes use of
incrementally augmenting the unlabeled data and using model
confidence.  Also like that work, we do not currently have a
theoretical framework for this (more complex) model.\footnote{Dasgupta
  et al. \shortcite{dasgupta01cotraining} proved, three years later,
  that a formal model roughly equivalent to the actual Blum and
  Mitchell algorithm \emph{does} have solid theoretical foundations.}
It would also be interesting to explore \emph{soft} hints, where the
range of $\chi$ is $[0,1]$ rather than $\{0,1\}$.

Recently, Ganchev et al.~\shortcite{ganchev08views} proposed a
co-regularization framework for learning across multiple related tasks
with different output spaces.  Their approach hinges on a constrained
EM framework and addresses a quite similar problem to that addressed
by this paper.  Chang et al.~\shortcite{chang08constraints} also
propose a ``semisupervised'' learning approach quite similar to our
own model.  The show very promising results in the context of semantic
role labeling.  Given the apparent (very!) recent interest in this
problem, it would be ideal to directly compare the different
approaches.

In addition to an analysis of the theoretical properties of the
algorithm presented, the most compelling avenue for future work is to
apply this framework to other task pairs.  With a little thought, one
can imagine formulating compatibility functions between tasks like
discourse parsing and summarization \cite{marcu-book00}, parsing and
word alignment, or summarization and information extraction.

\subsection*{Acknowledgments}

Many thanks to three anonymous reviewers of this papers whose
suggestions greatly improved the work and the presentation.  This work
was partially funded by NSF grant IIS 0712764.

\appendix
\section{Proofs} \label{sec:proofs}

The proof of Theorem~\ref{thm:pac} closes follows that of Blum and
Mitchell \shortcite{blum98cotraining}.

\begin{proof}[Proof (Theorem~\ref{thm:pac}, sketch)]
  Use the following notation: $c_k = \Pr_\cD[h(x)=k]$, $p_l =
  \Pr_\cD[f(x)=l]$, $q_{l|k} = \Pr_\cD[f(x)=l\|h(x)=k]$.  Denote by
  $\cA$ the set of outputs that satisfy the constraints.  We are
  interested in the probability that $h(x)$ is erroneous, given that
  $h(x)$ satisfies the constraints:

  \begin{small}
  \begin{align*}
    &\p{h(x) \in \cA\without\{l\} \| f(x)=l} \\
    &= \sum_{k\in \cA\without\{l\}} \p{h(x)=k \| f(x)=l} 
%    &= \sum_{k\in \cA\without\{l\}} \p{h(x)=k} \p{f(x) = l \| h(x) = k} / \p{f(x) = l} \\
    = \sum_{k\in \cA\without\{l\}} c_k q_{l|k} / p_l \\
    &\leq \sum_{k \in \cA} c_k (\card{\cY}-1 + \ep \sum_{l\neq k} 1/p_l) 
    \leq 2\sum_{k \in \cA} c_k (\card{\cY} - 1)
  \end{align*}
  \end{small}

  \noindent
  Here, the second step is Bayes' rule plus definitions, the third
  step is by the weak initial hypothesis assumption, and the last step
  is by algebra.  Thus, in order to get a probability of error at most
  $\eta$, we need $\sum_{k\in \cA} c_k = \Pr[h(x) \in \cA] \leq \eta /
  (2(\card{\cY}-1))$.
\end{proof}

\noindent
The proof of Corollary~\ref{cor:twosided} is straightforward.

\begin{proof}[Proof (Corollary~\ref{cor:twosided}, sketch)]
  Write out the probability of error as a double sum over true labels
  $y_1,y_2$ and predicted labels $\hat y_1,\hat y_2$ subject to
  $\chi(\hat y_1,\hat y_2)$.  Use the uncorrelation assumption and
  Bayes' to split this into the product two terms as in the proof of
  Theorem~\ref{thm:pac}.  Bound as before.
\end{proof}

\bibliography{bibfile}
\bibliographystyle{acl}

\end{document}